\documentclass[final]{cvpr}

\usepackage{times}
\usepackage{epsfig}
\usepackage{graphicx}
\usepackage{amsmath}
\usepackage{amssymb}

\usepackage[pagebackref=true,breaklinks=true,colorlinks,bookmarks=false]{hyperref}

\def\cvprPaperID{4755} 
\def\confYear{CVPR 2021}

\begin{document}

\title{Learning Position and Target Consistency for Memory-based Video Object Segmentation}


\author{Li Hu,  Peng Zhang,  Bang Zhang,  Pan Pan,  Yinghui Xu,  Rong Jin\\
Machine Intelligence Technology Lab, Alibaba Group\\
\tt\small \{hooks.hl, futian.zp, zhangbang.zb, panpan.pp, renji.xyh, jinrong.jr\}@alibaba-inc.com
}


\maketitle
\pagestyle{empty}
\thispagestyle{empty}

\begin{abstract}
This paper studies the problem of semi-supervised video object segmentation(VOS). Multiple works have shown that memory-based approaches can be effective for video object segmentation. They are mostly based on pixel-level matching, both spatially and temporally. The main shortcoming of memory-based approaches is that they do not take into account the sequential order among frames and do not exploit object-level knowledge from the target. To address this limitation, we propose to Learn position and target Consistency framework for Memory-based video object segmentation, termed as LCM. It applies the memory mechanism to retrieve pixels globally, and meanwhile learns position consistency for more reliable segmentation. The learned location response promotes a better discrimination between target and distractors. Besides, LCM introduces an object-level relationship from the target to maintain target consistency, making LCM more robust to error drifting. Experiments show that our LCM achieves state-of-the-art performance on both DAVIS and Youtube-VOS benchmark. And we rank the 1st in the DAVIS 2020 challenge semi-supervised VOS task.
\end{abstract}

\section{Introduction}

Video object segmentation(VOS) is a fundamental computer vision task, with a wide range of applications including video editing, video composition and autonomous driving. In this paper, we focus on the task of semi-supervised video object segmentation. Given a video and the ground truth object mask of the first frame, semi-supervised VOS predicts the segmentation masks of the objects specified by the ground truth mask in the first frame for the remaining frames. In video sequences, the target object will undergo large appearance changes due to continuous motion and variable camera view. And it may disappear in some frames due to occlusion between different objects. Furthermore, there are also similar instances of same categories that are difficult to distinguish, making the problem even harder. Therefore, semi-supervised VOS is extremely challenging despite the provided annotation in the first frame. 


\begin{figure}[!t]
\begin{center}
	\setlength{\fboxrule}{0pt}
	\fbox{\includegraphics[width=0.95\linewidth]{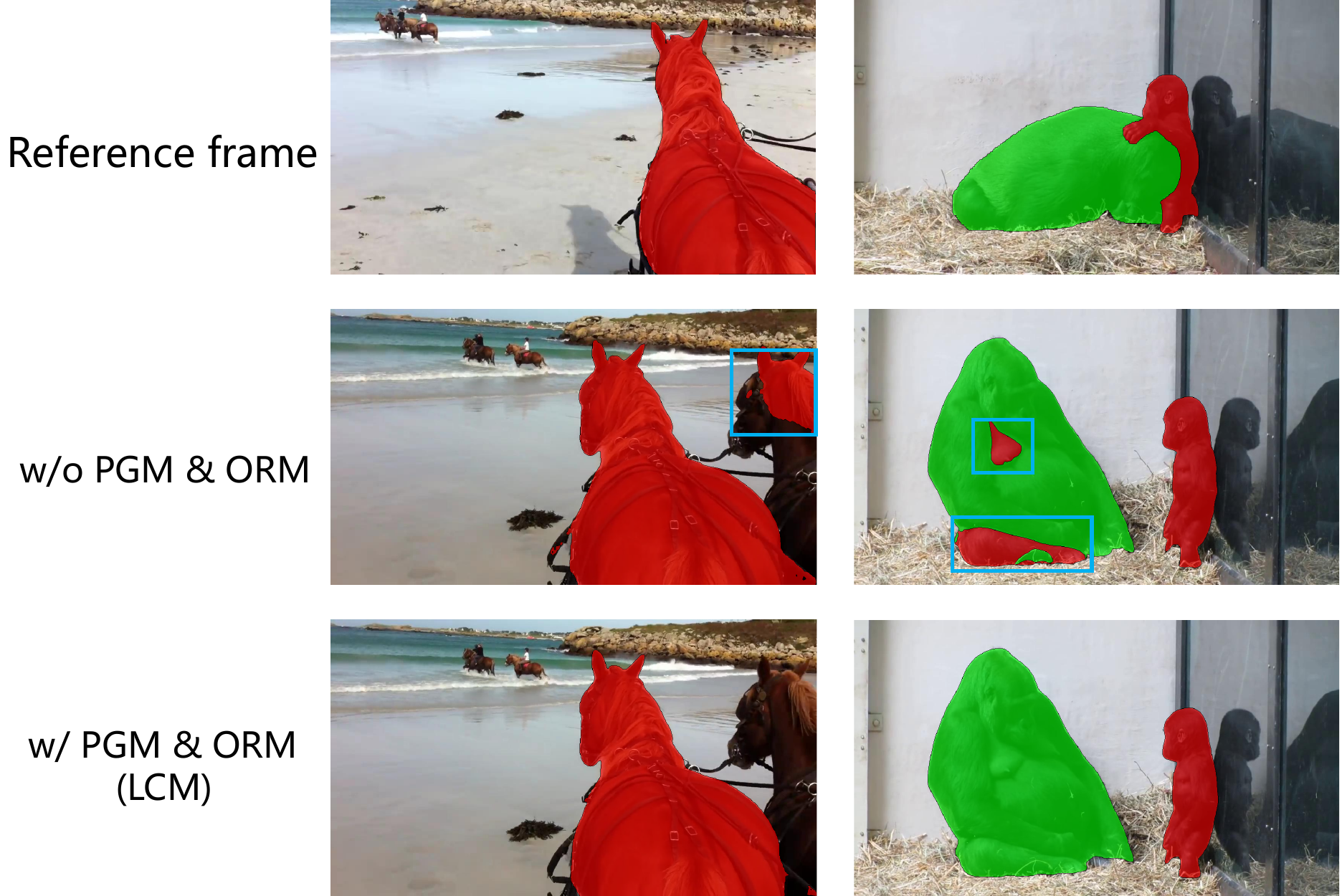}}
\end{center}
\caption{Typical memory-based approaches rely on pixel-level similarity, which leads to errors in prediction, as show in second row. The proposed Position Guidance Module(PGM) helps the network track the motion trajectory(bottom left).  And the object-aware Object Relation Module(ORM) prevents the network from making fragmented segmentation pieces(bottom right).}
\label{fig:fig1}
\end{figure}

The fundamental problem of VOS lies in how to make full use of the spatio-temporally structured information contained in video frames. Memory-based approaches are recently proposed with significant performance improvements in popular VOS benchmarks, e.g. DAVIS\cite{davis2016,davis2017} and Youtube-VOS\cite{youtubevos}. Space-Time Memory network(STM)\cite{STM} is the first memory-based semi-supervised VOS method, developing a memory mechanism to store information from all previous frames for the query frame to read. It differs from other matching-based methods as it expands its search range to the entire space-time domain and perform dense matching in the feature space. However, memory-based methods only consider pixel-level matching and tend to retrieve all pixels with high matching score in the query image. It may fail when a non-target region share similar visual appearance with the target regions
as illustrated in Figure~\ref{fig:fig1}. Recently, KMN\cite{KMN} introduces memory-to-query matching to improve STM. But the solution remains pixel similarity matching which cannot deal with appearance changes and deformation. In order to tackle the aforementioned issues, we propose to improve memory-based methods from two aspects: 1) Position consistency. 
The movement of objects usually follows a certain trajectory, which serves as an important instruction to guide segmentation. 2) Target consistency. 
The overall embedding feature for the tracked target should maintain object-level consistency throughout the entire video.


Propagation-based methods\cite{osmn,rgmp,agss-vos} introduce to directly utilize the prediction from previous frames for better segmentation. Inspired by these works, we propose to apply previous positional information as a guidance for memory-based methods to maintain position consistency. Typical matching-based methods\cite{videomatch, PML} only consider pixel-level feature without the context information from the entire object. Inspired by some works in tracking\cite{siamfc} and one/few-shot detection\cite{fsod,hsieh2019one}, we propose to integrate object-level feature into memory-based network to maintain target consistency. 

To this end, we propose a novel framework to Learn position and target Consistency for Memory-based video object segmentation(LCM).  
Taking advantage of STM, LCM performs pixel-level matching mechanism to retrieve target pixels based on similarity and stores previous information in a memory pool. This procedure is named Global Retrieval Module(GRM). Besides, LCM learns a local embedding named Position Guidance Module(PGM) to fully utilize the position consistency and guides the segmentation by learning a location response. To maintain target consistency, LCM introduces Object Relation Module(ORM). As the target object is annotated in the first frame of a video, the object relationship from the first value embedding is encoded to the query frame, which serves as a consistent fusion for context feature during the entire video sequence. Figure~\ref{fig:fig1} illustrates the effectiveness of our LCM against typical errors in memory-based methods.

Our contributions can be summarized as follows:
\begin{itemize}
\item We propose a novel Position Guidance Module to compute a location response to maintain position consistency in memory-based methods. 
\item We propose Object Relation Module to effectively fuse object-level information for maintaining consistency of the target object. 
\item We achieve state-of-the-art performance on both DAVIS and Youtube-VOS benchmark and rank the 1st in the DAVIS 2020 challenge semi-supervised VOS task.
\end{itemize}

\section{Related Works}
\textbf{Top-down methods for VOS. }
Top-down methods tackle video object segmentation with two processes. They first conduct detection methods to obtain proposals for target objects and then predict mask results. PReMVOS\cite{premvos} utilizes Mask RCNN\cite{maskrcnn} to generate coarse mask proposals and conducts refinement, optical flow and re-identification to achieve a high performance. DyeNet\cite{DyeNet} applies RPN\cite{fasterrcnn} to extract proposals and uses Re-ID Module to associate proposal with recurrent mask propagation. TAN-DTTM\cite{TAN-DTTM} proposes Temporal Aggregation Network and Dynamic Template Matching to combine RPN with videos and select correct RoIs. 
Top-down methods rely heavily on the pre-trained detectors and the pipelines are usually too complicated to conduct end-to-end training.

\textbf{Propagation-based methods for VOS. }
Propagation-based methods utilize the information from previous frames. MaskTrack\cite{MaskTrack} directly concatenates previous mask with current image as the input. RGMP\cite{rgmp} also concatenates previous masks and proposes a siamese encoder to utilize the first frame. OSMN\cite{osmn} designs a modulator to encode spatial and channel modulation parameters computed from previous results. AGSS-VOS\cite{agss-vos} uses current image and previous results and combines instance-specific branch and instance-agnostic branch with attention-guided decoder. 
In general, previous frame is similar in appearance to the current frame, but it cannot handle occlusion and error drifting. And previous works usually conduct implicit feature fusion which is lack of interpretability.

\textbf{Matching-based methods for VOS. }
Matching-based methods perform pixel-level matching between template frame and current frame.
PML\cite{PML} proposes a embedding network with triplet loss and nearest neighbor classifier. VideoMatch\cite{videomatch} conducts soft matching with foreground and background features to measure similarity. FEELVOS\cite{feelvos} proposes global and local matching according to the distance value. CFBI\cite{cfbi} applies background matching together with an instance-level attention mechanism. The main inspiration of our work is STM\cite{STM} which proposes to use all previous frames by storing information as memory. 
KMN\cite{KMN} applies Query-to-Memory matching to improve original STM with kernelized memory read.
Matching-based methods ignore the temporal information especially positional relationship. And they miss the knowledge from the overall target object. 

\begin{figure*}[!t]
\begin{center}
	\setlength{\fboxrule}{0pt}
	\fbox{\includegraphics[width=0.85\textwidth]{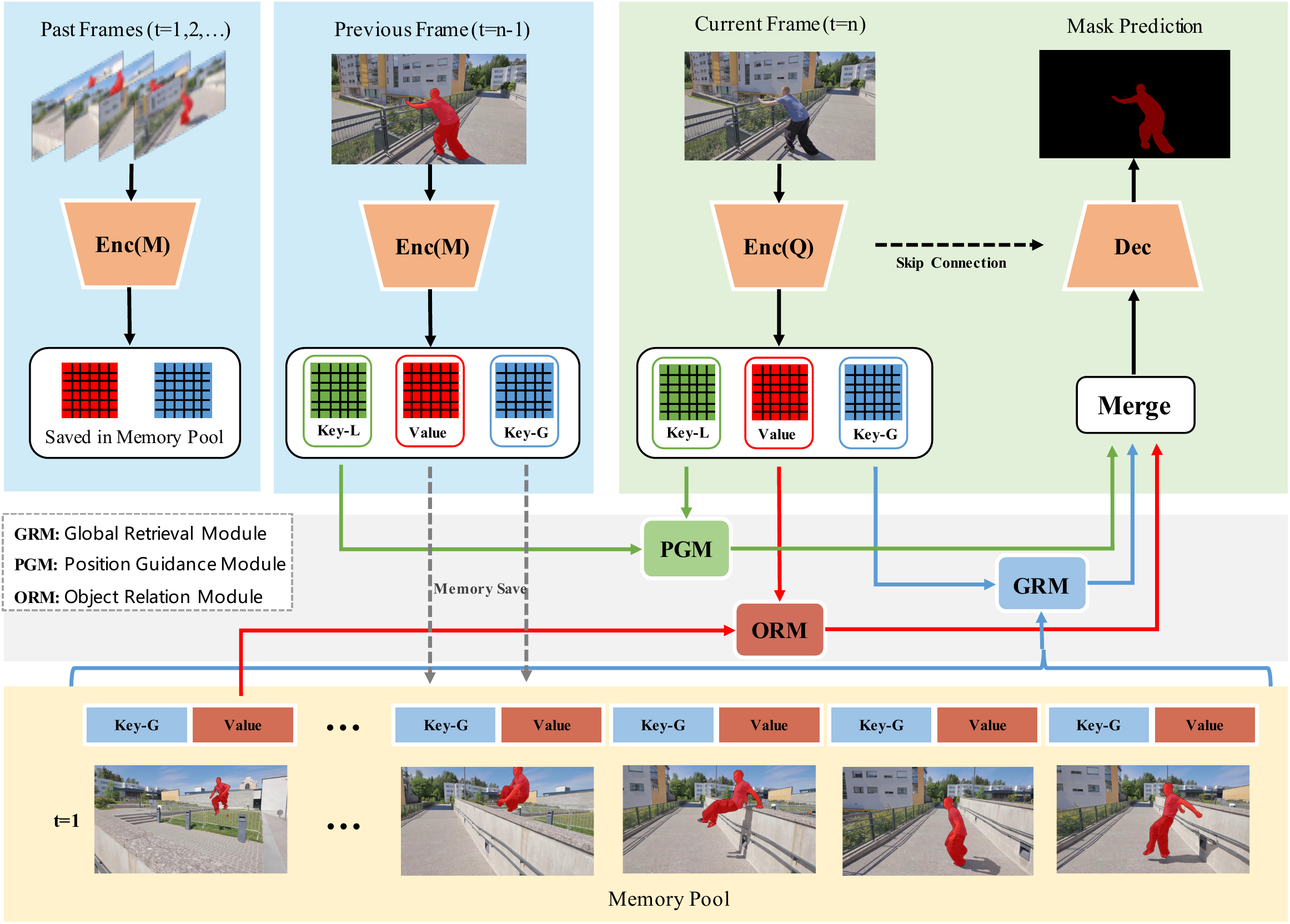}}
\end{center}
\caption{The overview of our LCM. The information of past frames are stored in memory pool. Global Retrieval Module(GRM) conducts pixel-level matching between query and memory pool. Position Guidance Module(PGM) encodes information from previous frame. Object Relation Module(ORM) fuses feature from first value embedding. }
\label{fig:overview}
\end{figure*}


\textbf{Attention mechanism.}
Attention is widely adopted in machine learning including natural language process and computer vision. 
Non-local\cite{nonlocal} network computes attention response at a position as a weighted sum of the features at all positions, capturing global and long-term information. \cite{empirical} proposes a generalized attention formulation for modeling spatial attention. Many semantic segmentation works\cite{ccnet,psanet,OCR} utilize attention to build context information for every pixels.
\cite{hsieh2019one} emphasizes the features of the query and images via co-attention and co-excitation.

\section{Methods}
We first present the overview of our LCM in section~\ref{sec:overview}. In section~\ref{sec:GRM}, we describe the Global Retrieval Module. Then we introduce the proposed Position Guidance Module and Object Relation Module in section~\ref{sec:PGM} and section~\ref{sec:PRM}. Finally, the detail of training strategy is in section~\ref{sec:train}.

\subsection{Overview}\label{sec:overview}
The overall architecture of LCM is illustrated in Figure~\ref{fig:overview}. LCM uses a typical Encoder-Decoder architecture to conduct segmentation. For a query image, the query encoder produces three embeddings, i.e. $Key$-$G$, $Key$-$L$ and $Value$. The embeddings are fully exploited in three sub-modules: Global Retrieval Module(GRM), Position Guidance Module(PGM) and Object Relation Module(ORM). First, GRM is designed the same as Space-Time Memory Network(STM)\cite{STM}. It calculates a pixel-level feature correlation between the current frame and memory pool. 
The $Key$-$G$ and $value$ from previous frames
are stored in memory pool via the memory encoder. Second, we propose PGM, which learns a feature embedding for both current frame and previous adjacent frame. Obviously previous frame contains similar positional information with current frame. Thus we build a positional relationship between these two frames which enhances positional constrain for the retrieved pixels. Moreover, to merge object-level information into pixel-level matching procedure and to prevent the accumulative error in memory pool, we propose ORM. The information of objects in the first frame will be maintained during entire sequential inference. Finally, we introduce the training strategy of our LCM. In the following section, we will further present a specific description.

\subsection{Global Retrieval Module}\label{sec:GRM}
Global Retrieval Module(GRM) highly borrows the implementation of Space-time Memory Network(STM)\cite{STM}. As illustrated in Figure~\ref{fig:overview}, Previous frames together with its mask predictions are encoded through the memory encoder meanwhile current frame is encoded through the query encoder. We use the ResNet-50\cite{resnet} as backbone for both encoders. For the $t$th frame, the output feature maps are defined as $r^{M}{\in}{\mathbb{R}}^{H{\times}W{\times}C}$ and $r^{Q}{\in}{\mathbb{R}}^{H{\times}W{\times}C}$. 
For previous frames,
the memory global key $k^{M}{\in}{\mathbb{R}}^{H{\times}W{\times}C/8}$ and memory value $v^{M}{\in}{\mathbb{R}}^{H{\times}W{\times}C/2}$ are embedded through two separated $3{\times}3$ convolutional layers from $r^{M}$. Then both embeddings are stored in memory pool and are concatenated along the temporal dimension, which are defined as $k^{M}_{p}{\in}{\mathbb{R}}^{T{\times}H{\times}W{\times}C/8}$ and $v^{M}_{p}{\in}{\mathbb{R}}^{T{\times}H{\times}W{\times}C/2}$. For query image, the query global key $k^{Q}{\in}{\mathbb{R}}^{H{\times}W{\times}C/8}$ will be embedded from $r^{Q}$. The Global Retrieval Module retrieves the matched pixel feature based on the similarity of the global key between query and memory pool by the following formulation:

\begin{equation}
s(i,j)=\frac{exp(k^{M}_{p}(i){\odot}{k^{Q}(j)^\mathsf{T}})}{\sum_{i}exp(k^{M}_{p}(i){\odot}{k^{Q}(j)^\mathsf{T}})}
\end{equation}
where $i$ and $j$ are the pixel feature indexes of memory pool and the query. $\odot$ represents the matrix inner production, and function $s$ denotes the $softmax$ operation, determining the location of the most similar pixel feature in memory pool for the query. Then the retrieved value feature is calculated as:
\begin{equation}
y^{GRM}(j)=\sum_{i}s(i,j){\odot}v^{M}_{p}(i)
\end{equation}

Global Retrieval Module encourages the query to search for the pixel-level appearance feature with high similarity along both spatial and temporal dimension. The main contribution of this module is its high recall. However, such mechanism does not fully utilize the characteristics of video object segmentation. The calculation of the correlation map is equally conducted with all features in memory pool without position consistency. As a consequence, the network tends to learn where to find the similar area but not correctly tracking the target object. The following proposed modules aim to solve above problems.

\subsection{Position Guidance Module}\label{sec:PGM}
In video object segmentation, the motion trajectory of an object is continuous and the recent frames usually contain the cues of approximate location of the target. When conducting Global Retrieval, all pixels with high similarity will be matched. Thus, if some small areas or other objects besides the tracked one have similar appearance feature, the Global Retrieval Module often incorrectly retrieves them as illustrated in Figure~\ref{fig:fig1}. Thus, the positional information from recent frames should be effectively used.

Here we introduce Position Guidance Module(PGM) which encodes previous adjacent frame to learn position consistency. As shown in Figure~\ref{fig:overview}, in addition to output global key, we also propose to extract local key from the $res4$ feature map for local position addressing. Specifically, another $3{\times}3$ convolutional layer is applied for both query embedding and previous adjacent memory embedding to output query local key $k^{Q}_{L}{\in}{\mathbb{R}}^{H{\times}W{\times}C/8}$ and memory local key $k^{M}_{L}{\in}{\mathbb{R}}^{H{\times}W{\times}C/8}$. 
\begin{figure}[!t]
\begin{center}
	\setlength{\fboxrule}{0pt}
	\fbox{\includegraphics[width=0.47\textwidth]{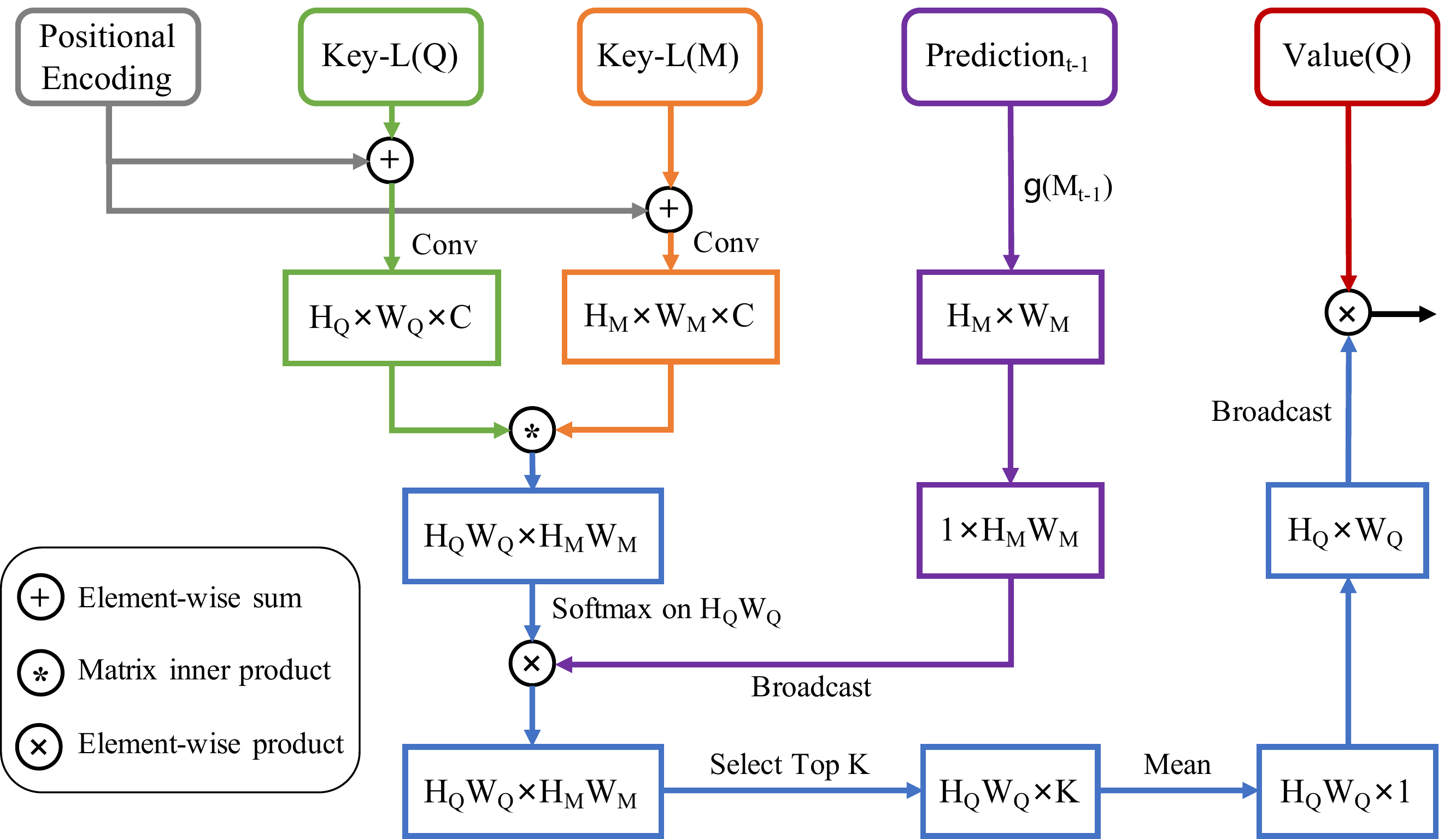}}
\end{center}
\caption{Implementation of Position Guidance Module.}
\label{fig:pos}
\end{figure}

The implementation of Position Guidance Module is depicted in Figure~\ref{fig:pos}. The global key is learned to encode visual semantics for matching robust to appearance variations as described in STM. In comparison, the local key is designed to not only address feature similarity but also encode positional correspondence. Since the matrix operation for these embeddings is position-invariant, we supplement them with 2D positional encodings\cite{imagetrans,detr} to maintain location cues. We use sine and cosine functions with different frequencies to define a fixed absolute encoding associated with the corresponding position, formulating it as $pos(i)$. Positional encodings are added to both local keys followed by a $1{\times}1$ convolutional layer $f_{n}$. We depict the process as follows.
\begin{equation}
p^{M}(i)=f_{n}(k^{M}_{L}(i)+pos(i))
\end{equation}
\begin{equation}
p^{Q}(j)=f_{n}(k^{Q}_{L}(j)+pos(i))
\end{equation}

Then we reshape $p^{M}$ and $p^{Q}$ and apply matrix inner product to get the embedding $S$ with size of $HW{\times}HW$. Softmax operation is applied on the query dimension to form a response distribution for each location in the previous frame. Meanwhile we use the previous predicted mask to reduce the response of non-object areas. The calculation is defined as:
\begin{equation}
S(i,j)=\frac{exp(p^{Q}(j){\odot}{p^{M}(i)^\mathsf{T}})}{\sum_{j}exp(p^{Q}(j){\odot}{p^{M}(i)^\mathsf{T}})}{\ast}g(M_{t-1})
\end{equation}
where $g(x)=\frac{exp(x)}{\emph{e}}$ prevents the response from the location of background close to zero since the previous prediction is not always correct. Next we select the top-K values on the memory dimension and average them to get the position map of size $H{\times}W$. Experimentally, we set $K=8$. The selected locations in the memory map determine a significant position association with corresponding query location. And the location with high response value in the position map represents the area where objects are most likely to appear in the query image. Finally, this position map serves as a spatial attention map and we conduct element-wise product between the position map and the query value $v^{Q}$:
\begin{equation}
y^{PGM}(j)=\frac{\sum_{i}{topK\{S(i,j)\}}}{K}{\ast}v^{Q}(j)
\end{equation}

\begin{figure}[!t]
\begin{center}
	\setlength{\fboxrule}{0pt}
	\fbox{\includegraphics[width=0.44\textwidth]{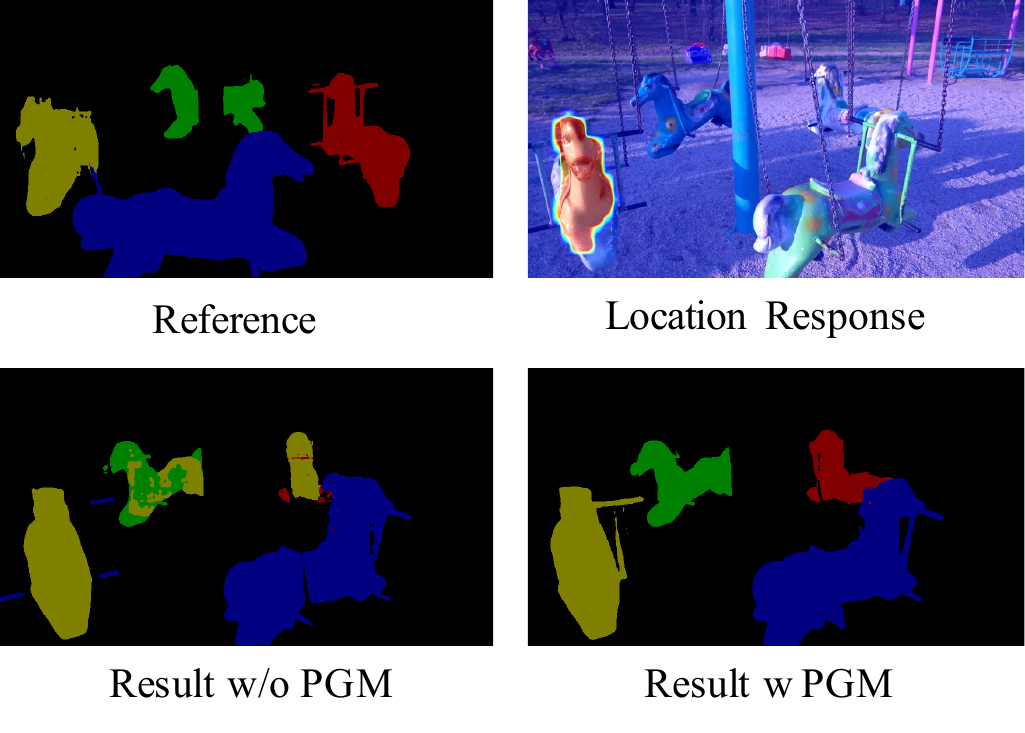}}
\end{center}
\vspace{-0.3cm}
\caption{The effectiveness of PGM.}
\label{fig:posv}
\end{figure}

To demonstrate the effectiveness of PGM, we illustrate the typical case in Figure~\ref{fig:posv}. Without PGM, pixels of similar objects are likely to be retrieved due to the high appearance similarity. As a comparison, PGM promotes a better discrimination between target and distractors.
We normalize the learned location response in PGM to a heatmap. The result shows that PGM learns a response distribution which not only considers the similarity of the appearance features between objects, but also correctly determines the location area of the target.


\subsection{Object Relation Module}\label{sec:PRM}
In video object segmentation, it is critical to utilize object-level feature of the target, which is not covered by above mechanism. The matching-based pixel retrieval is a bottom-up approach and lack of context information. During video inference, the accumulative error often brings noisy ($Key$-$G$, $value$) pairs into memory pool and will mislead the subsequent pixel matching process and position guidance as shown in middle right of Figure~\ref{fig:fig1}. To tackle above problems, it is essential to additionally utilize the first frame as it always provides intact and reliable masks. Specifically, we propose Object Relation Module(ORM) to fuse the object-level information of the first frame as a prior into the inference of entire video stream to maintain target consistency.

\begin{figure}[!t]
\begin{center}
	\setlength{\fboxrule}{0pt}
	\fbox{\includegraphics[width=0.47\textwidth]{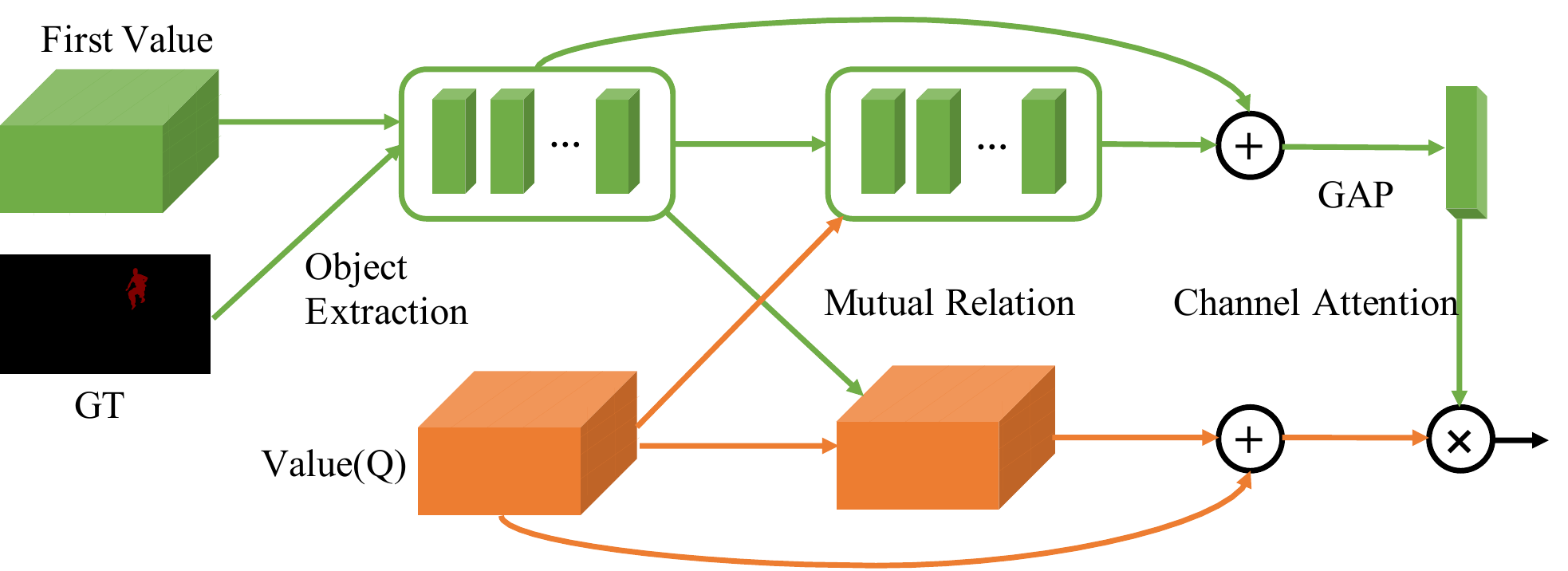}}
\end{center}
\caption{Process of Object Relation Module.}
\label{fig:prior}
\end{figure}

In Object Relation Module, we start from the first value $v^{F}$ and the query value $v^{Q}$. The module structure is illustrated in Figure~\ref{fig:prior}. According to the ground truth mask, for each object we select the foreground feature in the first value $v^{F}$ into a value set $F\{f_{i}\}$, where $i$ denotes the location that belongs to certain object mask. Inspired by \cite{hsieh2019one}, we design a cross relation mechanism to merge object-level feature into the query value. For both $F\{f_{i}\}$ and $v^{Q}(j)$, we conduct non-local operation and output respective non-local relation feature $F_Q\{f_{i}\}$ and $v_{F}^{Q}(j)$ as follows:
\begin{equation}
F_Q\{(f_{i})\}=\frac{1}{d}\sum_{j}{f(F\{f_{i}\},v^{Q}(j)){\ast}g(v^{Q}(j))}
\end{equation}
\begin{equation}
v_{F}^{Q}(j)=\frac{1}{d}\sum_{i}{f(v^{Q}(j),F\{f_{i}\}){\ast}g(F\{f_{i}\}})
\end{equation}
where $d=H{\ast}W$ is the normalization factor and $g$ is a $1{\times}1$ convolutional layer. $f$ denotes dot product between two vectors. Then the original feature is enhanced by the non-local relation feature via element-wise sum.
Furthermore, we conduct global average pooling on the enhanced first value feature followed by two fully-connected layers and Sigmoid function as in the design of SENet\cite{senet}, serving as the channel-wise attention. Thus, the query value can adaptively re-weighting the importance coefficient over channels through the instruction from object-level feature. The process is summarized as follows, where GAP indicates global average pooling:
\begin{equation}
v^{Q}(j) = v^{Q}(j)+v_{F}^{Q}(j)
\end{equation}
\begin{equation}
F\{f_{i}\} = F\{f_{i}\}+F_Q\{(f_{i})\}
\end{equation}
\begin{equation}
y^{ORM}(j) = v^{Q}(j){\ast}GAP(F\{f_{i}\})
\end{equation}

Object Relation Module encodes object-sensitive information flow into the feature extraction. The output is merged with Position Guidance Module and concatenated with the memory value from Global Retrieval Module as the final feature. We employ the decoder described in \cite{rgmp,STM} to gradually upsample the feature map combined with residual skip connections to estimate the object mask. We apply soft aggregation\cite{rgmp,STM} to merge the
multi-object predictions.

\subsection{Training Strategy}\label{sec:train}
\textbf{Pre-training on static images. }
As widely used in recent VOS task\cite{rgmp,STM,KMN}, we simulate fake video dataset with static images to pre-train the network for better parameter initialization. We leverage image segmentation datasets\cite{pre1,pre2,coco} for pre-training. A synthetic clip contains three frames. Specifically, one image is sampled from real dataset and generates other two fake images by applying random affine transforms. 

\textbf{Main-training on real videos without temporal limit. }
In this step, we leverage video object segmentation datasets to train the model. Different from the original main training setting in \cite{STM}, we do not limit video sampling intervals. Three frames are randomly selected from a video sequence and we randomly shuffle the order of them. 
Only objects that appear in all three frames are selected as foreground objects. 
This strategy encourages the network to strength retrieval capability since the target object will appear in all possible regions. 

\textbf{Fine-tuning on real videos as sequence. }
At inference of video object segmentation, the mask results is computed frame by frame sequentially. Therefore, in this training stage we further fine-tune the model to reduce the gap between training and testing. We sample three frames with time-order and the skip number is randomly selected from 1 to 5. The predicted soft mask result is used to compute memory embeddings. This training mechanism construct training samples with sequence information, which benefits the training of PGM.

\textbf{Training Details. }
We initialize the network with ImageNet pretrained parameters. During pre-training, we conduct translation, rotation, zooming and bluring to transform images and randomly crop 384$\times$384 patches. We minimize the cross-entropy loss using Adam optimizer with learning rate of 5e-4. During main-training and fine-tuning, we randomly crop a 640$\times$384 patch around the maximum bounding box of all objects in three frames. Adam optimizer with learning rate of 1e-5 is used in main-training and SGD optimizer with learning rate of 3e-4 for fine-tuning. We use 8 Tesla V100 GPUs.   Pre-training  takes  25  hours(10  epoch).   Training without temporal limit takes 12 hours(200 epoch). Training as sequence takes 3 hours(50 epoch). We do not apply post-processing or online training.


\section{Experiments}
We evaluate our model on DAVIS\cite{davis2016,davis2017} and YouTube-VOS\cite{youtubevos}, two popular VOS benchmarks with multiple objects. For YouTube-VOS, we train our model on the YouTube-VOS training set and report the result on YouTube-VOS 2018 validation set. For the evaluation on DAVIS, we train our model on DAVIS 2017 training set with 60 videos. Both DAVIS 2016 and 2017 are evaluated using an identical model trained on DAVIS 2017 for a fair comparison with the previous works. We also report the result trained with both DAVIS 2017 and YouTube-VOS(3471 videos) following recent works. 

The evaluation metric is the average of $\mathcal{J}$ score and $\mathcal{F}$ score. $\mathcal{J}$ score calculates the average IoU between the prediction and the ground truth mask. $\mathcal{F}$ score calculates an average boundary similarity between the boundary of the prediction and the ground truth mask.

\subsection{Compare with the State-of-the-art Methods}

\begin{table}
	\centering
	\begin{center}
\begin{tabular}{llllll} 
\hline
   &  &  \multicolumn{2}{c}{Seen} & \multicolumn{2}{c}{Unseen} \\  
   \cline{3-6}
   & Overall & $\mathcal{J}$ & $\mathcal{F}$ & $\mathcal{J}$ & $\mathcal{F}$ \\
\hline
OSMN\cite{osmn}            & 51.2 & 60.0 & 60.1 & 40.6 & 44.0 \\
MSK\cite{msk}              & 53.1 & 59.9 & 59.5 & 45.0 & 47.9 \\
RGMP\cite{rgmp}            & 53.8 & 59.5 & -    & 45.2 & -    \\
OnAVOS\cite{OnAVOS}        & 55.2 & 60.1 & 62.7 & 46.6 & 51.4 \\
RVOS\cite{rvos}            & 56.8 & 63.6 & 67.2 & 45.5 & 51.0 \\
OSVOS\cite{osvos}          & 58.8 & 59.8 & 60.5 & 54.2 & 60.7 \\
S2S\cite{s2s}              & 64.4 & 71.0 & 70.0 & 55.5 & 61.2 \\
A-GAME\cite{agame}         & 66.1 & 67.8 & -    & 60.8 & -    \\
PreMVOS\cite{premvos}      & 66.9 & 71.4 & 75.9 & 56.5 & 63.7 \\
BoLTVOS\cite{boltvos}      & 71.1 & 71.6 & -    & 64.3 & -    \\
DMM\cite{dmmnet}      & 58.0 & 60.3 & 63.5 & 50.6 & 57.4 \\
CapsuleVOS\cite{dmmnet} & 62.3 & 67.3 & 53.7 & 68.1 & 59.9 \\
GC\cite{GC}         & 73.2 & 72.6 & 75.6 & 68.9 & 75.7 \\
AFB\_URR\cite{AFB_URR}     & 79.6 & 78.8 & 83.1 & 74.1 & 82.6 \\
GraphMem\cite{GraphMem}    & 80.2 & 80.7 & 85.1 & 74.0 & 80.9 \\
CFBI\cite{cfbi}            & 81.4 & 81.1 & 85.8 & 75.3 & 83.4 \\
LWTL\cite{LWTL}            & 81.5 & 80.4 & 84.9 & \textbf{76.4} & \textbf{84.4} \\

KMN\cite{KMN}              & 81.4 & 81.4 & 85.6 & 75.3 & 83.3 \\

\hline 
STM\cite{STM}              & 79.4 & 79.7 & 84.2 & 72.8 & 80.9 \\
LCM                       & \textbf{82.0}   &  \textbf{82.2}  &  \textbf{86.7}  & 75.7   &  83.4  \\
\hline
\end{tabular}
\end{center}
	\caption{The quantitative evaluation on Youtube-VOS 2018 validation dataset.}
	\label{table:youtubevos}
\end{table}

\textbf{Youtube-VOS}\cite{youtubevos} is the largest dataset for video segmentation which consists of 4453 high-resolution videos. In detail, the dataset contains 3471 videos in the training set (65 categories), 474 videos in the validation set (additional 26 unseen categories). We train our model on Youtube-VOS training set and evaluate it on Youtube-VOS-18 validation set. 

As shown in Table~\ref{table:youtubevos}, our approach LCM obtains a final score of $82.0\%$, significantly outperforming our baseline STM($79.4\%$) of $2.6\%$. It demonstrates the effectiveness of our proposed modules on typical memory-based methods. Compared with other recent works, LCM also achieves state-of-the-art performance. CFBI\cite{cfbi} is built on a strong pipeline with COCO\cite{coco} pre-trained DeepLabV3+\cite{v3+} and a well-designed segmentation head. KMN applies a Hide-and-Seek training strategy which improves the diversity and accuracy of training data and is a general data-augmentation for any other memory-based VOS methods including LCM.  Without these enhancements, our performance is still higher. 
This result demonstrates the robustness and generalization of our approach on a complex dataset.

\begin{table}
	\centering
	\begin{center}
\begin{tabular}{lcccc} 

\hline
  & $\mathcal{J}$ Mean & $\mathcal{F}$ Mean & Overall \\
\hline
 \multicolumn{4}{c}{Validation Set} \\
\hline
OSVOS\cite{osvos}  & 56.6 &  63.9 &  60.3 \\
PReMVOS\cite{premvos}  & 73.9 & 81.7 & 77.8 \\
OSVOS$^s$\cite{osvos_s}  &  64.7 &  71.3 &  68.0 \\
OSMN\cite{osmn}   & 52.5 &  57.1 & 54.8   \\
VideoMatch\cite{videomatch}   & 56.5  &  68.2 & 62.4   \\

RGMP\cite{rgmp}   & 64.8 & 68.6 &  66.7 \\
A-Game\cite{agame}  &  67.2 & 72.7 & 70.0 \\
FAVOS\cite{favos}  &  54.6 &  61.8 &  58.2 \\
FEELVOS\cite{feelvos}(+YV)     & 69.1 &  74.0 &  71.5 \\
DMM\cite{dmmnet}     &  68.1 &  73.3 &  70.7 \\
RANet\cite{ranet}      &  63.2 &  - &  65.7 \\ 

GC\cite{GC}  & 69.3 & 73.5 & 71.4 \\
AFB\_URR\cite{AFB_URR}  & 73.0 & 76.1 &  74.6 \\
LWTL\cite{LWTL}(+YV)   &  79.1  & 84.1 & 81.6 \\
CFBI\cite{cfbi}(+YV)   &  79.1 &  84.6 &  81.9 \\

GraphMem\cite{GraphMem}(+YV)  & 80.2 & 85.2 & 82.8 \\
KMN\cite{KMN}(+YV)    &  80.0 & 85.6 &  82.8 \\

\hline
STM\cite{STM}    & 69.2  &  74.0  &  71.6   \\
LCM       & 73.1  &  77.2  & 75.2 \\
\hline
STM\cite{STM}(+YV)   & 79.2 & 84.3 &  81.8 \\ 
LCM(+YV)      & \textbf{80.5} &  \textbf{86.5} &  \textbf{83.5} \\
\hline
 \multicolumn{4}{c}{Test-dev Set} \\
\hline
PReMVOS\cite{premvos}  & 67.5 & 75.7 & 71.6 \\
RGMP\cite{rgmp}   & 51.3 & 54.4 & 52.9 \\
FEELVOS\cite{feelvos}(+YV)   &  55.2 & 60.5 &  57.8 \\
RANet\cite{ranet}      &  53.4  & -    & 55.3 \\

CFBI\cite{cfbi}(+YV)  & 71.1 &  78.5 &  74.8 \\
KMN\cite{KMN}(+YV)     &  74.1 & 80.3 & 77.2 \\

\hline
STM\cite{STM}(+YV)  &  69.3  &  75.2 & 72.2  \\
LCM(+YV)      & \textbf{74.4} &  \textbf{81.8} &  \textbf{78.1} \\
\hline
\end{tabular}
\end{center}
	\caption{The quantitative evaluation on DAVIS-2017 validation and test-dev dataset. (+YV) indicates training with both DAVIS and Youtube-VOS.  }
	\label{table:davis2017}
\end{table}

\textbf{DAVIS 2017}\cite{davis2017} is a multi-object extension of DAVIS 2016 and it is more challenging than DAVIS 2016 since the model needs to consider the difference between various objects. The validation set of DAVIS 2017 consists of 59 objects in 30 videos. In this section we evaluate our model on both DAVIS 2017 validation and test-dev benchmark. 

The results are compared to state-of-the-art approaches in Table~\ref{table:davis2017}. 
Our method shows state-of-the-art results. When applying both DAVIS and Youtube-VOS datasets for training, LCM achieves $83.5\%$, surpassing our baseline STM of $1.7\%$. And LCM also shows higher performance than other existing methods including online-learning methods and offline-learning methods. Follwing recent work, we also report the result with only DAVIS for training. And LCM outperforms the baseline STM of $3.6\%$.
In addition, we report the result on the DAVIS testing split and also shows best results of $78.1\%$, surpassing STM by a significant margin($+5.9$). By employing similar approaches in LCM together with other tricks such as better backbone, strong segmentation head, multi-scale testing and model ensemble, we achieve $84.1\%$ on the DAVIS challenge split and rank the 1st in the DAVIS 2020 challenge semi-supervised VOS task.

\textbf{DAVIS 2016}\cite{davis2016} consists of 20 videos annotated with high-quality masks each for a single target object. As shown in Table~\ref{table:davis2016}, LCM also achieves state-of-the-art performance. Compared to other methods, LCM is slightly higher than KMN of $0.2\%$. Since DAVIS 2016 is relatively a simple dataset and its performance highly relies on the precision of segmentation detail. A possible reason is that the Hide-and-Seek can provide more precise boundaries as described in KMN. Compared to the baseline STM, LCM shows better accuracy ($89.3$vs.$90.7$).

We also report the running time on DAVIS2016. We use 1 Tesla P100 GPU for inference. The increased running time brought by PGM and ORM is no more than 6\% compared with the baseline STM. We also compare it with other existing methods and our LCM maintains a comparable fast inference speed with higher performance.

\begin{table}
	\centering
	
\begin{center}
\begin{tabular}{lcccc} 
\hline
 & Time(s) & $\mathcal{J}$ Mean & $\mathcal{F}$ Mean & Overall \\
\hline
OSVOS\cite{osvos} & 9 & 79.8 & 80.6 & 80.2 \\
MaskRNN\cite{maskrnn} & - & 80.7 & 80.9 & 80.8 \\  
LSE\cite{LSE} & - & 82.9 & 80.3 &  81.6 \\
CINN\cite{cinn} & 30 & 83.4 & 85.0 & 84.2 \\
PReMVOS\cite{premvos} & 32.8 & 84.9 & 88.6 & 86.8 \\
OnAVOS\cite{OnAVOS} & 13 & 86.1 & 84.9 &  85.5 \\
RANet\cite{ranet}   & 4 &  86.6 &  87.6 & 87.1 \\
FEELVOS\cite{feelvos} & 0.45   & 81.1 &  82.2 &  81.7 \\
RGMP\cite{rgmp} & 0.13 & 81.5 & 82.0 &  81.8 \\
A-Game\cite{agame} & 0.07 &  82.0 & 82.2 & 82.1 \\
FAVOS\cite{favos} & 1.8 &  82.4 &  79.5 &  81.0 \\
DMVOS\cite{dmvos} & 0.035 & 87.8 & 87.5 & 88.0\\
RANet\cite{ranet}   & 0.03 &  85.5 & 85.4 & 85.5 \\ 
GC\cite{GC}   & 0.04 &  87.6 & 85.7 &  86.6 \\
CFBI\cite{cfbi} & 0.18 &  88.3 & 90.5 &  89.4 \\

KMN\cite{KMN}  & 0.12 &  89.5 & \textbf{91.5} & 90.5 \\

\hline
STM\cite{STM} & 0.112 & 88.7 & 89.9 &  89.3 \\ 
LCM   & 0.118 & \textbf{89.9} & 91.4 & \textbf{90.7}  \\
\hline
\end{tabular}
\end{center}
	\caption{The quantitative evaluation on DAVIS-2016 validation dataset. The running time of STM is our reimplement result.}
	\label{table:davis2016}
\end{table}

\begin{figure*}[h]
\begin{center}
	\setlength{\fboxrule}{0pt}
	\fbox{\includegraphics[width=0.95\linewidth]{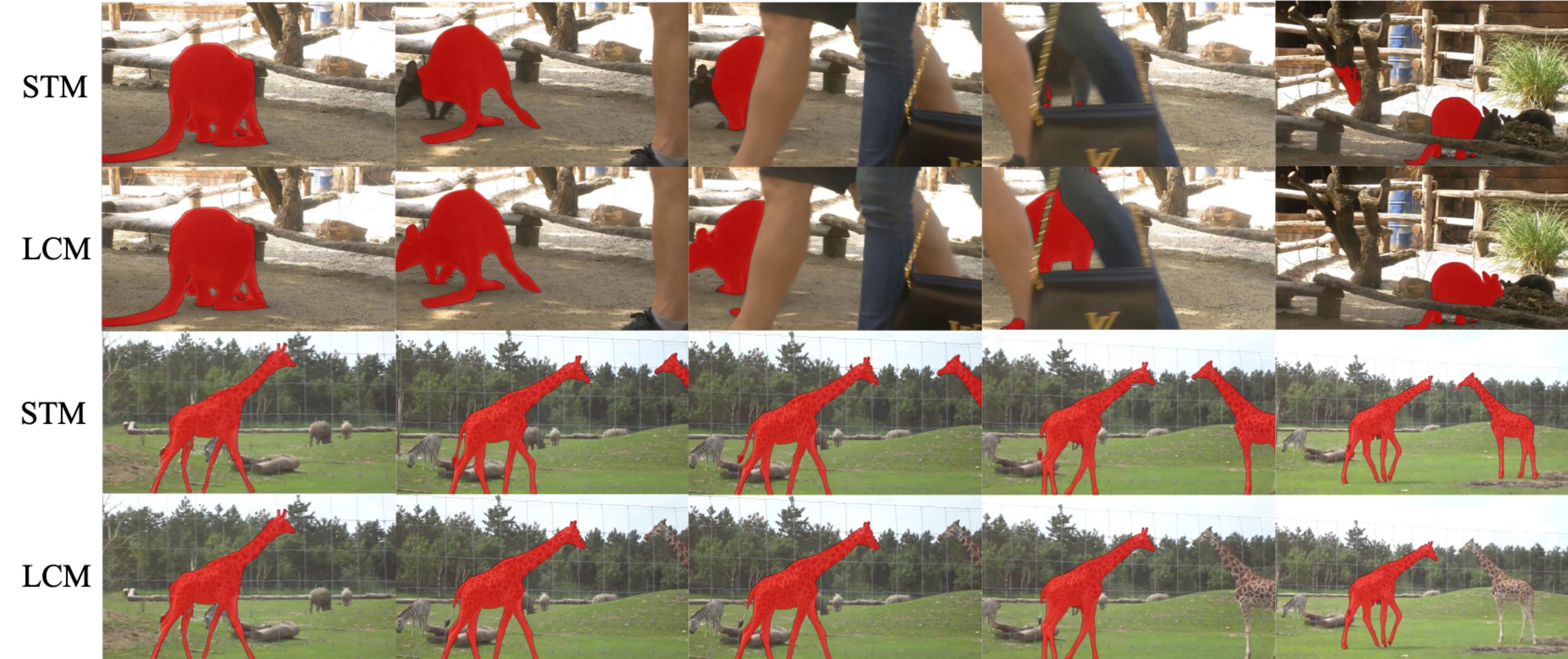}}
\end{center}
\caption{Qualitative results of our proposed LCM. Our model is more robust under challenging situation such as occlusion, appearance change and similar objects.}
\label{fig:viz1}
\end{figure*}

\subsection{Qualitative Results. }
We show the qualitative results compared with memory-based method STM in Figure~\ref{fig:viz1}. We use the author’s officially released pre-computed results. 
The result shows that LCM can reduce typical errors in memory-based method and is more robust under challenging situation such as occlusion, appearance change and similar objects.

\subsection{Ablation Study}
We conduct an ablation study on DAVIS 2017 validation set to demonstrate the effectiveness of our approach.

\textbf{Network Sub-module. }
We experimentally analyze the effectiveness of our proposed three sub-modules. In this experiment, we do not apply pre-training step for saving time and directly use DAVIS and Youtube-VOS to train our model. The result is shown in Table~\ref{table:ablation1}. When applying all three proposed modules, LCM achieves 79.2$\%$ on DAVIS 2017 validation set without pre-training. The performance drops to 77.8$\%$ and 78.4$\%$ respectively When we disable Position Guidance Module or Object Relation Module. Without both modules, the result degrades to 76.9$\%$, which demonstrates the importance of these two modules. Furthermore, when disabling Global Retrieval Module, the performance heavily drops from 79.2$\%$ to 67.5$\%$. The reason is that Global Retrieval Module is the fundamental module of LCM otherwise a large amount of information is absent without memory pool. 

\textbf{Training Strategy. }
We experimentally analyze the impact of our training strategy. The result is shown in Table~\ref{table:ablation2}. When only conducting pre-training and training without temporal limit, the performance achieves 82.9$\%$, which is already a state-of-the-art performance. When only conducting pre-training and training as sequence, the result degrades to 80.7$\%$.The reason is that small sampling interval makes the model incapable to learn appearance change and fast motion. Consequently, our framework has the best performance when combining all three training stages.

\section{Conclusion}
This paper investigates the problem of memory-based video object segmentation(VOS) and proposes Learning position and target Consistency of Memory-based video object segmentation(LCM). We follow memory mechanism and introduce Global Retrieval Module(GRM) to conduct pixel-level matching. Moreover, we design Position Guidance Module(PGM) for learning position consistency. And we integrate object-level information with Object Relation Module(ORM). Our approach achieves state-of-the-art performance on VOS benchmark.

\begin{table}
	\centering
	\begin{center}
\begin{tabular}{ccccccc} 

\hline
 GRM & PGM & ORM & $\mathcal{J}$ Mean & $\mathcal{F}$ Mean & Overall \\
 \hline
  $\checkmark$ & $\checkmark$ & $\checkmark$ & 77.1 & 81.4 & 79.2 \\
 \hline
  $\checkmark$ &  & $\checkmark$ & 75.5 & 80.1 & 77.8\\
  $\checkmark$ & $\checkmark$ &  & 76.0 & 80.8 & 78.4\\
  $\checkmark$ &  &  & 74.6 & 79.2 & 76.9\\
 & $\checkmark$ & $\checkmark$ & 65.2 & 69.8 & 67.5\\
 
\hline
\end{tabular}
\end{center}
	\caption{Ablation study of the network sub-module on DAVIS 2017 validation without pre-training.}
	\label{table:ablation1}
\end{table}

\begin{table}
	\centering
	\begin{center}
\begin{tabular}{lcccc} 

\hline
Training Strategy & $\mathcal{J}$  & $\mathcal{F}$  & Avg \\
 \hline
Combining three training stages & 80.5 & 86.5 & 83.5 \\
 \hline
 w/o training as sequence  & 79.9 & 85.9 & 82.9\\
w/o training without temporal limit  & 77.9 & 83.5 & 80.7\\
 
\hline
\end{tabular}
\end{center}
	\caption{Ablation study of the training strategy on DAVIS 2017 validation.}
	\label{table:ablation2}
\end{table}


{\small

\bibliographystyle{IEEEtran}
}

\end{document}